\documentclass[11pt]{article}
\usepackage[a4paper,margin=2.45cm]{geometry}
\usepackage[T1]{fontenc}
\usepackage[utf8]{inputenc}
\usepackage{lmodern}
\usepackage{microtype}
\usepackage{amsmath,amssymb}
\usepackage{booktabs,tabularx,array}
\usepackage{graphicx}
\usepackage{caption}
\usepackage{enumitem}
\usepackage{url}
\usepackage[hidelinks]{hyperref}
\hypersetup{pdftitle={When Zero Is Competitive: A Descriptive Pilot of Transformer and Gate-Graph Surrogates on Random-Angle Quantum Circuits},pdfauthor={Mahesh Reddy Pagadala},pdfsubject={Descriptive pilot on quantum-circuit observable surrogates},pdfkeywords={quantum circuits,neural surrogates,negative results,zero baseline}}
\setlist{nosep,leftmargin=*}
\title{When Zero Is Competitive: A Descriptive Pilot of\\Transformer and Gate-Graph Surrogates on Random-Angle Quantum Circuits}
\author{Mahesh Reddy Pagadala\\Aston University, Birmingham, UK\\\texttt{mahesh15116@gmail.com}}
\date{November 2026}
\begin{document}
\maketitle

\begin{abstract}
We evaluate supervised classical prediction of all single-qubit $Z_i$ and pairwise $Z_iZ_j$ expectation values for noiseless, layered RX/RY/RZ--CNOT circuits with 4--10 qubits. Across 63 balanced workload cells and disjoint circuit-template evaluation, we compare one transformer-encoder instantiation and gate-dependency graph neural networks (GateDAGs) against literal-zero and training-derived constant predictors. In the Arm~K training labels, 44.89\% of scalar targets have absolute magnitude below 0.05. The equal-instance-weighted held-out-template mean absolute error (MAE) of literal zero is 0.193399; no evaluated transformer checkpoint or final GateDAG endpoint improves upon it. A separate product-state RY control attains MAE 0.007755 versus 0.514811 for zero, demonstrating learning on an easier analytic mapping but not explaining the entangled-workload result. A training-fitted mean constant improves upon zero by only 0.14\% on the Arm~K \emph{training} mean-squared-error measure, a descriptive point estimate with no uncertainty interval. The tested transformer's 815,233-parameter source identity is verified locally, but its packed input-field semantics have not been independently audited against the recovered source. This is an exploratory descriptive pilot, not an impossibility result, formal equivalence test, or identification of target concentration as the unique cause. An independently implemented generator can produce a comparable workload from the specified design; exact frozen data and model-run evidence remain private.
\end{abstract}

\section{Introduction}
Classical surrogates aim to predict quantum-circuit observables from circuit structure and continuous parameters, potentially avoiding repeated state-vector simulation. For certain ensembles, however, a low aggregate prediction error does not by itself imply learning a useful circuit-dependent map. If many expectation values are near zero, a constant estimator may obtain a competitive aggregate score even without reading circuit inputs.

We therefore ask a narrow empirical question: do the particular transformer and GateDAG implementations evaluated here improve held-out-template prediction of raw local $Z/ZZ$ observables over elementary constants? The workload was stratified by qubit count, round count and connectivity family; angles were drawn independently, with both fresh-angle evaluation on seen templates and disjoint-template evaluation. The study does \emph{not} test an additional workspace-token intervention and does not assert that arbitrary neural surrogates fail.

Our contribution is a transparent \emph{descriptive pilot} that (i) reports the target-magnitude distribution and explicit constant comparators, (ii) evaluates two architectural instantiations under documented data/exposure choices, (iii) distinguishes pooled-scalar and equal-instance weighting, and (iv) reports a restricted analytic positive control and an entangled fresh-angle check. This pilot does not establish any general sample-complexity boundary or causal explanation of the model results.

\section{Related work and positioning}
Cantori and Pilati~\cite{cantori2025} study supervised prediction of magnetisation for circuits with single-qubit rotations interleaved with CNOTs, finding sharp differences between interlayer and interqubit angle-variation regimes. This precedent rules out a claim that P002 first discovers that random-angle circuit outputs can be challenging. Their structured angles, global magnetisation target and MSE/$R^2$ metrics are different from our independent RX/RY/RZ angles, multiple CNOT families, all raw $Z_i$ and $Z_iZ_j$ targets, and MAE-versus-constant analysis. Our distribution is qualitatively closer to unrestricted variation across layers but is not a matched test of their variance-controlled learning boundary.

Liu \emph{et al.}~\cite{liu2026} report successful output regression with attention-based circuit GNNs in settings that include H/T/CNOT circuits, selected/rescaled targets, and noisy and noiseless conditions. Our GateDAG instead uses continuous rotations and recurrent gate-wire message passing, outputs all signed one- and two-qubit observables, and evaluates on disjoint generated templates. These joint differences in gate set, target representation, graph model, split and noise assumptions preclude a direct performance comparison. The present negative endpoints neither reproduce nor contradict their positive results. Liao \emph{et al.}~\cite{liao2026} report efficient predictive surrogates for specified quantum-processor objectives; this does not imply a learnability guarantee for our uniformly sampled angles.

Neural quantum states~\cite{carleo2017,carleo2018}, classical shadows~\cite{huang2020}, classical quantum-property learning~\cite{huang2022many,huang2022process} and separations for observable learning~\cite{molteni2026} concern related but distinct information settings. Barren-plateau theory for parameterised quantum models~\cite{mcclean2018,nemkov2025} must not be equated with optimisation of the \emph{classical} neural networks tested here. The importance of informative negative results and negative-control baselines is emphasised in related methodological discussions~\cite{karl2024,petersen2025}. Other classical-surrogate formulations~\cite{schreiber2023} and surrogate-based error mitigation~\cite{liao2026mitigation} likewise do not define the present experiment.

\section{Workload, splits and metrics}
\subsection{Circuit ensemble and labels}
Each circuit starts in $\lvert0\rangle^{\otimes n}$ with $n\in\{4,\ldots,10\}$, round count $r\in\{2,4,6\}$, and connectivity family $f\in\{\mathrm{path},\mathrm{ring},\mathrm{all\_to\_all}\}$. These choices define $7\times3\times3=63$ balanced cells. Each round applies one independently axis-selected gate from $\{\mathrm{RX},\mathrm{RY},\mathrm{RZ}\}$ on each qubit in ascending index order, followed by $n$ CNOT gates. The CNOT endpoints are sampled from allowed edges of the selected family, with randomly chosen control direction; each rotation angle is independently $U[-\pi,\pi)$. Thus a circuit has $2nr$ elementary gates.

For each angle instance we compute the noiseless state-vector expectation values
\begin{equation}
  y(C,\boldsymbol\theta)
    =\bigl(\langle Z_1\rangle,\ldots,\langle Z_n\rangle,
           \langle Z_1Z_2\rangle,\ldots,\langle Z_{n-1}Z_n\rangle\bigr),
  \qquad Q(n)=\frac{n(n+1)}{2}\in[10,55].
\end{equation}
The labels were generated by an archived custom NumPy double-precision state-vector simulator, not by Qiskit, CUDA-Q or finite-shot hardware measurement. Distinct fixed NumPy PCG64 streams generate topology, training angles, within-template angles and held-out-template angles. The original generator source hashes and RNG states are recorded in the frozen manifest; changing the sampler can create a \emph{comparable} workload without recreating the identical frozen dataset.

\subsection{Templates and sampling panels}
A \emph{template} fixes gate order, rotation axes, qubit indices, and CNOT endpoints/directions, but not numerical rotation angles. Each cell contains 30 training templates and two distinct held-out templates, for 2,016 templates total. Each training template receives 20 independent training-angle instances and four fresh within-template angle instances drawn from the \emph{same} $U[-\pi,\pi)$ distribution, with independent RNG streams. Each held-out template receives four instances. The frozen 45,864 examples are split into 37,800 training, 7,560 within-template fresh-angle, and 504 held-out-template examples. The held-out panel therefore spans 126 distinct template identities (two per cell), not 504 independently drawn template identities.

The Arm~K training subset comprises 30 templates per cell and four training angles per template: $63\times30\times4=7,560$ examples. A model fitted on Arm~K uses 7,560 within-template examples and the common 504-example held-out panel for evaluation. The within-template panel measures new-angle prediction on seen gate sequences, whereas the held-out panel tests different gate sequences.

\subsection{Metrics and simple estimators}
The training loss is masked, equal-instance mean squared error (MSE):
\begin{equation}
  \mathcal L_{\mathrm{MSE}}=\frac1N\sum_{a=1}^{N}\frac{1}{Q(n_a)}
  \sum_{q=1}^{Q(n_a)}\bigl(\widehat y_{a,q}-y_{a,q}\bigr)^2.
\end{equation}
The primary reported prediction metric replaces the squared residual with its absolute value, still averaging per example before averaging examples. Consequently it is \emph{not} equivalent to pooling every target scalar across variable qubit counts. We compare the literal-zero estimator with train-derived per-$n$, per-observable mean and median constants, fitted exclusively from Arm~K training labels.

Across 226,800 scalar labels in the Arm~K training set, the mean absolute value is 0.168849, the median absolute value is 0.064561, 44.89\% have $|y|<0.05$, 19.22\% have $|y|>0.3$, and 10.48\% have $|y|>0.5$. These are properties of the target distribution, \emph{not} evidence that the circuit-input function is intrinsically unlearnable.

\begin{table}[t]
\centering\small
\caption{Pooled-scalar MAE for the Arm~K panels, reported separately from the paper's equal-instance primary metric. Constants were fitted only on training labels.}
\label{tab:scalar}
\begin{tabular}{lrrr}\toprule
Estimator & Train & Within-template & Held-out templates\\\midrule
Literal zero & 0.168849 & 0.169009 & 0.170763\\
Train-fitted mean & 0.169606 & 0.169906 & 0.171786\\
Train-fitted median & 0.168833 & 0.169028 & 0.170789\\\bottomrule
\end{tabular}
\end{table}
On the same Arm~K held-out panel, literal zero's \emph{equal-instance} MAE is 0.193399. In particular, 0.170763 (pooled-scalar) and 0.193399 (equal-instance) must not be substituted for one another. Close performance of these three sensible constant estimators does not imply that every possible constant is competitive.

\section{Models and training protocol}
\subsection{Transformer encoder}
The transformer has a 128-dimensional hidden representation, four encoder layers, four attention heads and a 512-dimensional feedforward sublayer; a CPU-only model reconstruction counted \textbf{815,233 trainable parameters} for the $W=0$ setting. This is the archived P001 model with four checked literal dimension substitutions from the original source (width $64\to128$, feedforward width $128\to512$, layers $3\to4$, and a corresponding gather-width expansion $64\to128$). Both original and substituted source SHA-256 values matched the logged P002 run identity. Every evaluated arm uses the baseline with no learned non-input workspace tokens ($W=0$); no $W=0$ versus $W=4$ workspace-prediction study was carried out.

The archived protocol describes intended gate-type, ordered qubit-index, role, position and qubit-count embeddings, sine/cosine angle features and dedicated observable queries. The recovered executable \texttt{pack()} produces five integer arrays of shape $(179,)$, a float32 array of shape $(179,2)$, and three integer scalar fields. \emph{Only the shapes, types, source hashes and parameter count have been locally verified;} the mapping of each packed field to intended token semantics has not been independently checked against the recovered implementation. The model's negative performance is therefore explicitly restricted to this particular implementation, rather than a general assertion about transformers.

The training script uses AdamW (learning rate $10^{-3}$, weight decay 0.01), batch size 32, per-instance masked MSE, gradient clipping at norm 1, fixed update budgets and evaluations every 64 updates; no learning-rate scheduler or early stopping is implemented. Seeds are 101, 202 and 303 in every arm.
\begin{table}[t]
\centering\small
\caption{Exploratory transformer data/exposure arms. All rows use the same $d=128$, four-layer, four-head, FFN-512, $W=0$ architecture.}
\label{tab:arms}
\begin{tabular}{lrrrr}\toprule
Arm & Templates/cell & Angles/template & Train instances & Updates\\\midrule
B & 6 & 4 & 1,512 & 384\\
A & 6 & 20 & 7,560 & 384\\
K & 30 & 4 & 7,560 & 384\\
L & 6 & 4 & 1,512 & 1,920\\\bottomrule
\end{tabular}
\end{table}

\subsection{GateDAG and restricted controls}
The GateDAG uses gate-kind and qubit-identity embeddings, sine/cosine rotation-angle and positional features, followed by recurrent predecessor-message propagation along gate-wire dependencies, GRU cells and an observable query readout. Depth configurations of 12, 24 and 40 message-passing rounds have 823,809, 1,604,865 and 2,646,273 trainable parameters respectively. The optimiser, masked loss, batch size and clipping follow the same basic conventions as above, with longer 7,680-update depth experiments at seed 202. These are single-seed exploratory runs.

A separate four-qubit, one-layer product-state RY control uses analytical targets $\langle Z_i\rangle=\cos\theta_i$ and $\langle Z_iZ_j\rangle=\cos\theta_i\cos\theta_j$, with 2,048 training and 512 held-out-angle examples, seed 202 and 1,000 updates. It is deliberately easier than the entangled main workload. Another diagnostic trains on 64 entangled instances from 16 $n=4,r=2$ path-drawn templates and evaluates 64 additional fresh-angle examples from the same templates after 1,500 updates.

\section{Results}
\subsection{Held-out template MAE}
Across all four transformer arms and three seeds, none of the \emph{recorded evaluated checkpoints} has held-out equal-instance MAE below the literal-zero value 0.193399. The best observed transformer score in the archived checkpoints is approximately 0.194460; this is a \emph{selected exploratory checkpoint}, not a fixed-final-endpoint performance guarantee. In this paper, non-improvement means only that the reported model point estimate does not fall below the zero estimator on the same panel. No pre-registered practical-effect threshold, significance test or equivalence margin is implied.

\begin{table}[t]
\centering\small
\caption{GateDAG endpoint scores, with the same held-out zero MAE of 0.193399 for all rows. These runs are not independent seeds.}
\label{tab:gnn}
\begin{tabular}{lrrrr}\toprule
Regime & Parameters & Updates & Train MAE & Held-out MAE\\\midrule
K, 12 rounds & 823,809 & 1,920 & 0.190717 & 0.193497\\
K, 12 rounds & 823,809 & 7,680 & 0.190476 & 0.193707\\
K, 24 rounds & 1,604,865 & 7,680 & 0.191281 & 0.194118\\
K, 40 rounds & 2,646,273 & 7,680 & 0.191269 & 0.194155\\\bottomrule
\end{tabular}
\end{table}
\begin{figure}[t]
\centering
\includegraphics[width=\linewidth]{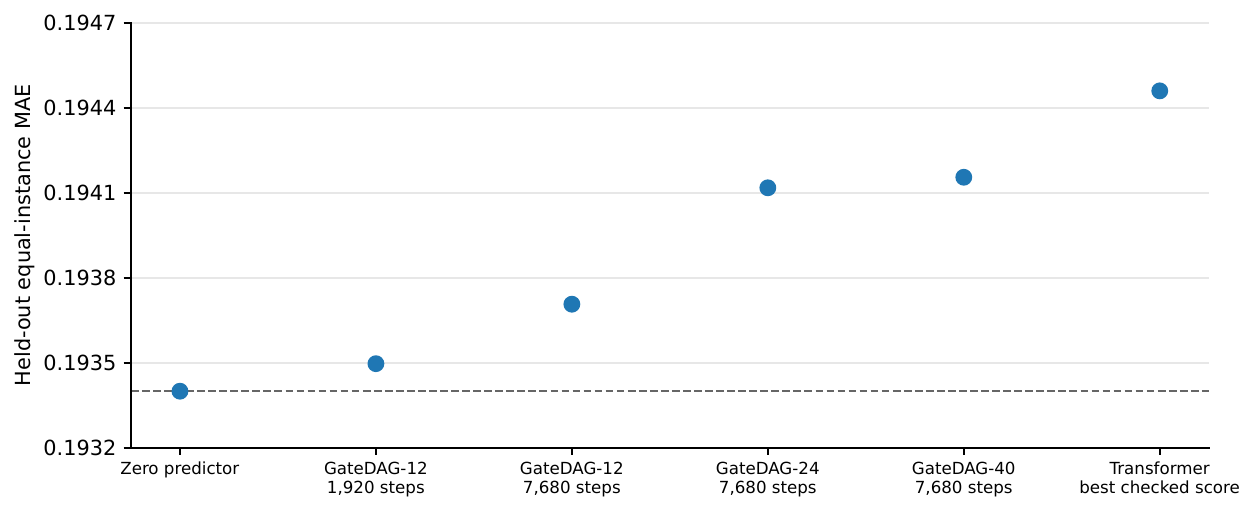}
\caption{Descriptive held-out equal-instance MAE. The zero-predictor level is dashed. GateDAG points are the stated run endpoints; the transformer point is the best \emph{evaluated} checkpoint among 12 exploratory runs. The y-axis shows held-out equal-instance MAE (0.1932--0.1947); this narrow range is intentionally displayed to show small differences. No error bars or uncertainty inference are available for the single-seed GateDAG sweep.}
\label{fig:mae}
\end{figure}

No final GateDAG endpoint in Table~\ref{tab:gnn} improves on zero. In the extended 12-round trajectory, however, the checkpoint at step 7,040 had held-out MAE 0.1933898901, beating zero 0.1933994673 by approximately $9.58\times10^{-6}$. This transient near-tie is not evidence of a reliable gain in the absence of seed-level variation or paired uncertainty intervals. Increasing message-passing depth and the tested update budget did not yield a useful endpoint improvement, but this cannot rule out alternative representations, skip connections, optimisation schedules or training regimes. A graph-structure audit found the median longest gate predecessor path 19, the 90th percentile 30, and maximum 40; merely matching or exceeding that path length does not guarantee trainability.

In Arm~L (1,512 training examples), final training MAE fell to about 0.122, but within-template and held-out errors were about 0.226 and 0.234, respectively. Fitting the restricted training set did not establish generalisation to fresh angles or templates.

\subsection{Positive control and limited entangled fitting}
On the product-state RY control, GateDAG achieved held-out-angle MAE 0.007755, compared with 0.514811 for the zero predictor (Z-only 0.006748; ZZ-only 0.008426). This establishes that the training pipeline can learn this restricted analytic map, but it does not certify the representation or trainability of the entangled mapping.

On the 64-instance entangled training diagnostic, MAE after 1,500 updates was 0.171882 versus zero 0.402068. The fresh-angle panel instead had MAE 0.425408 versus zero 0.414512. Thus some in-sample fitting is visible, but neither complete memorisation nor reliable fresh-angle prediction was established by this small diagnostic.

\subsection{MSE audit and compute feasibility}
\begin{table}[t]
\centering\small
\caption{Frozen-label baseline MSE under \emph{equal-instance} weighting for Arm~K. These are baseline metrics, not trained-model fixed-checkpoint MSE scores.}
\label{tab:mse}
\begin{tabular}{lrrr}\toprule
Estimator & Train & Within-template & Held-out templates\\\midrule
Literal zero & 0.095369 & 0.094946 & 0.097297\\
Train-fitted mean constant & 0.095236 & 0.094958 & 0.097276\\\bottomrule
\end{tabular}
\end{table}
The Arm~K training-MSE difference between train-fitted mean and literal zero is about 0.14\% of the zero baseline, a descriptive point estimate with no uncertainty interval. Across the last 512 logged minibatches, the mean \emph{online training} MSE values were 0.094992 (D12), 0.095266 (D24) and 0.095261 (D40). Because these losses are measured on changing minibatches with changing model weights, they cannot be used as a fixed final-model or held-out-model MSE comparison. The final GNN weights were not archived, and a formal model-versus-constant equivalence test is not available. The 10-step CUDA resource preflight completed at depth 40 with approximately 954 MiB peak reserved memory on the reported 8 GB GPU, showing that device-memory failure did not block these configurations.

\section{Discussion}
For this particular independently uniform-angle RX/RY/RZ--CNOT distribution, many raw signed $Z/ZZ$ observables are close to zero, and inexpensive fitted constants are competitive on the recorded error measures. The relevant lesson is not that zero is privileged or globally Bayes-optimal; it is that a learned predictor should be judged against multiple sensible constant estimators before low absolute error is interpreted as input-dependent skill. The task is deterministic given full circuit input and does not contain an irreducible measurement-noise floor merely because targets vary.

Multiple mechanisms remain compatible with these measurements. The representation or observable query head may discard predictive information, deep graph message passing may be hard to optimise, the training loss may encourage near-constant predictions under the observed label distribution, or the evaluated data and optimisation budgets may be inadequate. The product-state control is a basic learning-pipeline check on a different task; the restricted entangled fit reveals a train-versus-fresh-angle gap. Neither identifies target concentration as a \emph{cause}, isolates one of those mechanisms, or proves intrinsic unlearnability. P002 deliberately did not perturb target concentration while holding all other aspects of the learning problem fixed.

A better general reporting practice is to publish target-magnitude distributions, literal-zero and training-fitted constants, and both scalar-pooled and equal-instance metrics when output dimensionality varies. Conditional errors for large $|y|$ would distinguish failures on the near-zero bulk from partial learning in the tails. That diagnostic is recommended here but was not archived for the trained models; the manuscript makes no model-specific high-magnitude-tail claim.

\section{Limitations and scope}
This pilot covers only 4--10-qubit finite-depth noiseless simulations with the specified gate, connectivity and parameter distributions. Transformer arms use three training seeds, while the detailed GateDAG depth sweep uses a single seed. The experiments have no preregistered equivalence margin or paired neural-minus-constant uncertainty intervals, and the transient near-tie cannot establish a reliable gain. No comparison of wall-clock speed with simulation, quantum hardware data, or statistical confirmation of architecture-family failure was performed.

Conditional trained-model MAE and MSE restricted to targets with $|y|>0.3$ or $|y|>0.5$ were not preserved and cannot be reconstructed from aggregate endpoint logs. Thus models might have learned something in the high-magnitude tail despite their aggregate results. Trained-model full-panel held-out MSE and archived model weights are unavailable; logged online MSE is not an equivalent replacement. The transformer's packed feature \emph{shapes} and source identity are known, but the implementation-level field semantics have not been independently audited. A $W=0$--$W=4$ workspace intervention was not tested and is deferred to a separate study.

\section{Data, code and reproducibility}
The repository \url{https://github.com/lonewolf15116/quantum-workspace-study} is \emph{private} at the time of this manuscript. The author reports a private local SHA-256-inventoried archive, \texttt{p002\_reproducibility\_v3\_private.zip}, containing 110 files, with digest
\begin{center}\small\ttfamily\detokenize{de0dc6d99c24b25fc710d82ac2eea8ce02a8a715432ea51281015e3d96fe7be4}\end{center}
It contains the frozen P002 dataset, manifests, training/evaluation logs, positive-control evidence, scripts and archived P001 simulator/model sources. The archive is \emph{not} downloadable from the repository. Its integrity was checked locally, not independently by the present manuscript reviewer.

A local read-only paper-evidence verifier returned \texttt{PASS\_COMPLETE} with 24 programmed checks, zero missing items and zero reported errors; Appendix~\ref{app:checks} enumerates their scope. A separate audit checked generator and simulator source hashes against the frozen manifest and reproduced 12 sampled label vectors exactly using the archived custom NumPy \texttt{simulate()} function (maximum absolute difference 0.0) in the author's local environment. This is sampled numerical regeneration, not exhaustive regeneration of 45,864 examples or an independent rerun of model training. Another read-only CPU audit matched archived original and dimension-substituted transformer source hashes to the experimental identity and counted 815,233 parameters. For the legacy D12 run, the verifier accepted an otherwise matching identity lacking an explicit depth field using the 823,809 parameter count; this does not prove byte-identical source provenance.

A competent reader can implement an \emph{independently generated comparable ensemble} from the circuit, sampling and split rules in Section~3. This does \emph{not} recover the same 2,016 template identities, 45,864 angle instances, train/test run outputs or source revisions. Exact frozen-experiment replication requires author-approved access to private data and run artifacts, and access is not guaranteed. This study therefore does not claim a publicly reproducible benchmark. The current environment snapshot included in the archive is not a complete historical dependency lockfile.

\section{Conclusion}
On a frozen layered random-angle CNOT workload with raw local $Z/ZZ$ targets, the evaluated transformer and GateDAG configurations did not deliver a lower reported held-out-template equal-instance MAE than the literal-zero baseline at the compared checkpoints and endpoints. Because zero and sensible fitted constants are competitive, baseline-aware evaluation is essential for interpreting model error. The limited positive control, single-seed GateDAG depth sweep and unavailable conditional tail errors prevent causal, inferential or architecture-family conclusions. These findings are presented solely as a transparent descriptive pilot.

\appendix
\section{Inventory of the 24 local paper-evidence checks}\label{app:checks}
The read-only \texttt{verify\_p002\_paper\_evidence.py} checks the presence/hash of selected files and the following dataset and logged metric properties. Numerical comparisons use an absolute tolerance of $10^{-6}$, except exact structural conditions. These are 24 local consistency checks, \emph{not} 24 independent training replications:
\begin{enumerate}[label=\arabic*.]
\item Dataset cardinalities (2,016 templates; 45,864 instances; panel counts).
\item Template-identity/panel split consistency.
\item D12 legacy identity metadata (missing explicit layers key; parameters 823,809).
\item D12 training MAE at step 7,680.
\item D12 held-out MAE at step 7,680.
\item D12 7,680-step training log completeness.
\item D24 identity metadata.
\item D24 training MAE at step 7,680.
\item D24 held-out MAE at step 7,680.
\item D24 7,680-step training log completeness.
\item D40 identity metadata.
\item D40 training MAE at step 7,680.
\item D40 held-out MAE at step 7,680.
\item D40 7,680-step training log completeness.
\item Arm~K training scalar mean $|y|$.
\item Arm~K fraction of scalars with $|y|<0.05$.
\item Held-out fitted-mean-constant pooled-scalar MAE.
\item Arm~K training zero MSE.
\item Held-out zero MSE.
\item D12 last-512-step online training-minibatch MSE.
\item D24 last-512-step online training-minibatch MSE.
\item D40 last-512-step online training-minibatch MSE.
\item Restricted entangled diagnostic training MAE.
\item Restricted entangled diagnostic within-template fresh-angle MAE.
\end{enumerate}
Checks accept one documented legacy metadata exception and make no claims that training MSE is a fixed held-out model score. The separate 12-label regeneration and 110-member archive-integrity checks are distinct, limited audit procedures rather than interchangeable evidence of full scientific reproduction.

\end{document}